%% file: acl_latex.tex
\pdfoutput=1

\documentclass[11pt]{article}

\usepackage{acl}

\usepackage{times}
\usepackage{latexsym}

\usepackage[T1]{fontenc}

\usepackage[utf8]{inputenc}

\usepackage{microtype}

\usepackage{inconsolata}

\usepackage{graphicx}
\usepackage{amsmath} 

\title{Explaining Veracity Predictions with Evidence Summarization: A Multi-Task Model Approach}

\author{Recep Firat Cekinel \\
  Middle East Technical University \\
  \texttt{rfcekinel@ceng.metu.edu.tr} \\\And
  Pinar Karagoz \\
  Middle East Technical University \\
  \texttt{karagoz@ceng.metu.edu.tr} \\}

\begin{document}
\maketitle
\begin{abstract}
The rapid dissemination of misinformation through social media increased the importance of automated fact-checking.
Furthermore, studies on what deep neural models pay attention to when making predictions have increased in recent years. While significant progress has been made in this field, it has not yet reached a level of reasoning comparable to human reasoning. To address these gaps, we propose a multi-task explainable neural model for misinformation detection. Specifically, this work formulates an explanation generation process of the model's veracity prediction as a text summarization problem. Additionally, the performance of the proposed model is discussed on publicly available datasets and the findings are evaluated with related studies. 
\end{abstract}

\input{sections/1-intro}

\input{sections/2-relatedworks}
\input{sections/3-method}
\input{sections/4-results}

\input{sections/5-conclusion}

\section{Limitations}

First, the T5 and Flan T5 models were pre-trained massively on English corpora. Consequently, the performance of these models on languages with limited resources may not be satisfactory. Secondly, the validation experiments revealed significant fluctuations in the model's performance when utilizing certain hyperparameter sets. Therefore, the hyperparameter optimization was a critical part of the evaluation process. Furthermore, the interpretability of the generated explanations may vary depending on the complexity of the text.  Therefore, future research should address these limitations to enhance the robustness and applicability of our approach.

\section*{Acknowledgements}
We gratefully acknowledge the computational resources kindly provided by METU Robotics and Artificial Intelligence Center.

\bibliography{anthology,custom}

\appendix
\input{appendix}

\end{document}

%% file: sections/1-intro.tex
\section{Introduction}

Fake news is considered as media content that contains misinformation and can mislead people \cite{shu2017fake, zhou2020survey}. Advancements in social networking and social media not only facilitate information accessibility but also cause the rapid spread of fake news on social media \cite{vosoughi2018spread}. Consequently, fake news becomes a powerful tool for manipulating public opinion, as observed during influential events like the 2016 US Presidential Election and the Brexit referendum \cite{pogue2017stamp, allcott2017social}. To address this issue, automated fake news detection methods have emerged, aiming to determine the veracity of claims while minimizing human effort \cite{oshikawa2020survey}.




Multi-task learning (MTL) is a technique in machine learning to train similar tasks at the same time by leveraging their differences and commonalities \cite{crawshaw2020multi, chen2021multi, zhang2021survey}. Additionally, MTL allows data utilization as the model can transfer knowledge between tasks. Notably, the insights gained while learning one task can benefit other related tasks, leading to better generalization across tasks. Moreover, from the business point of view, deploying a single multi-task model may reduce the complexity of maintenance and resource requirements.

This paper primarily focuses on designing a multi-task explainable misinformation detection model. To be more specific, a fact-checking model is trained on veracity prediction and text summarization tasks simultaneously. The generated summaries are derived from evidence documents and serve as justifications for the model's veracity prediction. Therefore, it should not be considered as a post-hoc explainability model. The contribution of the work lies in the use of multi-task learning for fact-checking and text summarization together, particularly through a new architecture including different neural models. The tasks, fact-checking and summarization, complement each other such that one does misinformation detection while the other explains the reason for the model's decision. The source codes are available at: \url{https://github.com/firatcekinel/Multi-task-Fact-checking}

%% file: sections/2-relatedworks.tex
\section{Related Work}

Automated fake news detection studies have been studied from data mining \cite{shu2017fake} and natural language processing \cite{oshikawa2020survey, guo2022survey, vladika2023scientific} perspectives. \cite{zhou2020survey} classify the previous studies into four groups: knowledge-based \cite{ pan2018content, cui2020deterrent}, style-based \cite{zhou2020fake, perez2018automatic, jin2016novel, jwa2019exbake}, propagation-based \cite{hartmann2019mapping, zhou2019network}, and source-based \cite{sitaula2020credibility}. 

Kotonya and Toni \cite{kotonya2020explainablesurvey} present a survey on explainable fact-checking that categorized the studies based on their methods for generating explanations. These methods include exploiting neural network artifacts \cite{popat2017truth, popat2018declare, shu2019defend, lu2020gcan, silva2021propagation2vec}, rule-based approaches \cite{szczepanski2021new, gad2019exfakt, ahmadi2020rulehub}, summary generation \cite{atanasova2020generating, kotonya2020explainable, stammbach2020fever, brand2022neural}, adversarial text generation \cite{thorne2019evaluating, atanasova2020adversarial, dai2022ask}, causal inference for counterfactual explanations \cite{cheng2021causal, zhang2022causalrd, li2023boosts, xu-etal-2023-counterfactual}, neurosymbolic reasoning \cite{pan-etal-2023-fact} and Q\&A \cite{ousidhoum-etal-2022-varifocal, yang2022explainable}.


The most related study in the literature was the E-BART model \cite{brand2022neural} that was trained for both classification and summarization by introducing a joint prediction head on top of the BART \cite{lewis-etal-2020-bart} language model. In other words, the encoder and decoder of the BART model are shared for both tasks. In contrast to this approach, this work incorporates the T5 Encoder as a shared module. For summarization, a T5 Decoder is trained while feed-forward layers are employed for classification. We also measured the effect of using two loss weighting strategies and evaluated the impact of instruction fine-tuning by switching the T5 model with the Flan-T5 \cite{chung2022scaling} version.

%% file: sections/3-method.tex
\section{Method}

In this study, a multi-task model that is based on the T5 \cite{raffel2020exploring} transformer is proposed. The model is trained on text summarization and veracity prediction tasks jointly. T5 transformer is an encoder-decoder model that converts each task to a text-to-text problem. Flan-T5 \cite{chung2022scaling} model employs instructional fine-tuning to further improve the T5 model that is also utilized in the evaluation.

\begin{figure} [t]
    \centering
    \includegraphics[scale=0.3]{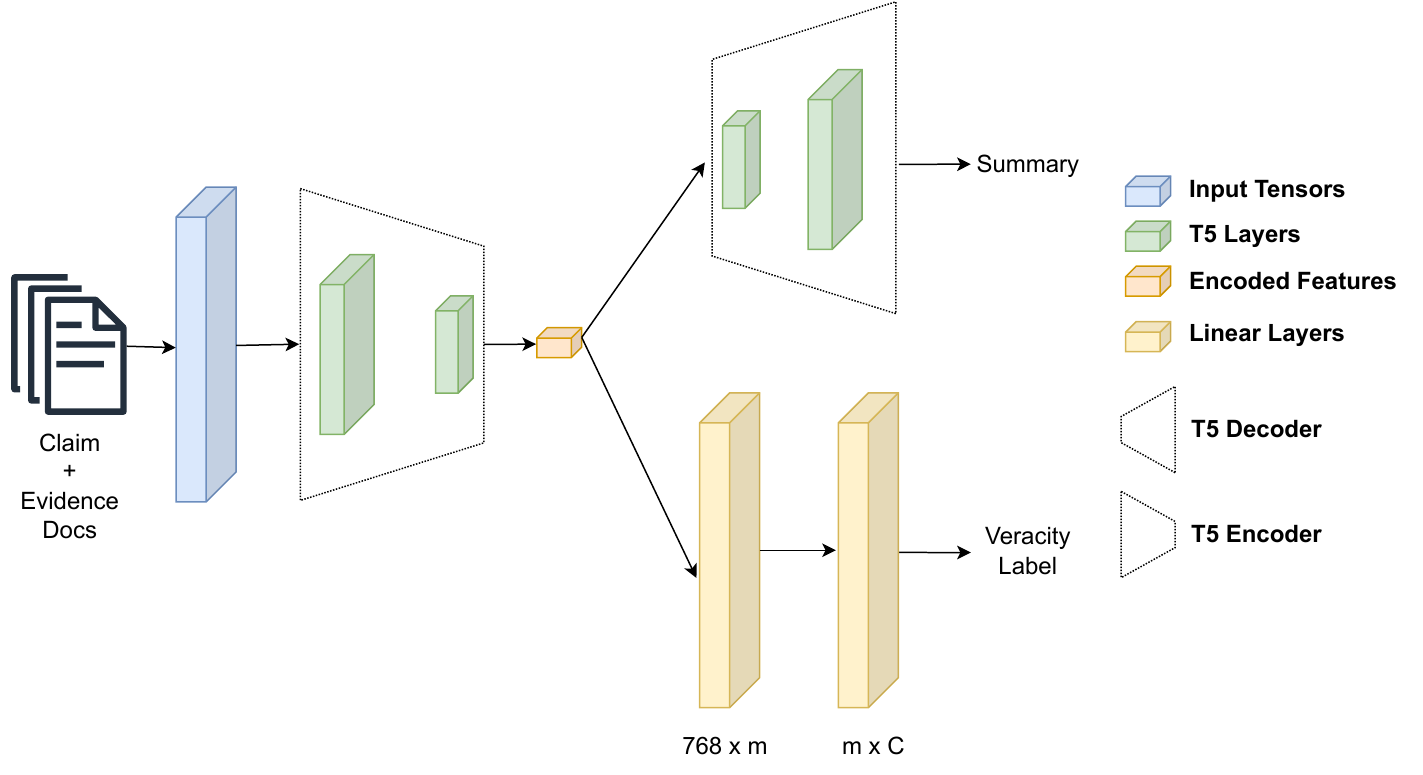}
    \caption{The multi-task model architecture}
    \label{fig:t5_mt}
\end{figure}

The model architecture is given in Figure \ref{fig:t5_mt}. Both summarization and classification tasks share a T5 Encoder during training. At first, the T5 Encoder encodes the claim and evidence sentences in a latent space. Afterwards, the T5 Decoder produces a summary using the T5 Encoder's representation. Simultaneously, for the veracity prediction, the encoder's output is processed by two feed-forward layers respectively. We employ the ReLU activation function and apply dropout between two linear layers and the sigmoid activation function after the second linear layer. Besides, the cross entropy loss is used for measuring summary and classification losses. 

The overall loss is calculated by taking the weighted sums of the summary loss ($w_{s}$) and the classification loss ($w_{c}$) that is given in Equation \ref{eq:static-loss}. 
\begin{equation}
    Loss = w_{s}*Loss_{summ} + w_{c}*Loss_{cl} \label{eq:static-loss}
\end{equation}

Two loss weighting strategies are employed: i) static loss coefficients and ii) uncertainty weighting. For the static loss coefficients, constant weights are set for the classification and summarization losses prior to training. To determine the optimal weights, grid search-based validation experiments are performed. In addition to the static loss coefficients, this paper also utilizes the uncertainty weighting strategy \cite{kendall2018multi} that enables dynamic adjustment of the weights based on prediction confidence. Subsequently, the overall loss is calculated by taking the weighted sums of the summary loss and the classification loss.  

\begin{figure*} [t]
    \centering
    \includegraphics[scale=0.85]{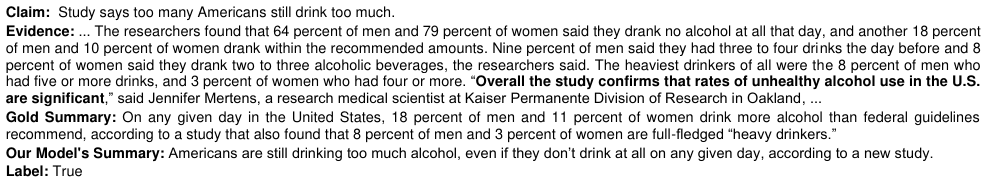}
    \caption{A sample claim from PUBHEALTH \cite{kotonya2020explainable} with our model's outputs}
    \label{fig:sample-claim}
\end{figure*}

Figure \ref{fig:sample-claim} presents an example claim alongside our model's predictions. Based on supplementary information provided under the "Evidence" section, the claim has been verified by a reviewer. The gold standard summary was also authored by human annotators, while the abstractive summary was generated by a T5-based multi-task model. The generated summary not only aligns with the veracity label but can be considered as an explanation of the model's reasoning behind its decision. More examples are provided in Appendix \ref{sec:summary-appendix}.

%% file: sections/4-results.tex
\section{Experimental Results}

In this section, the proposed model was evaluated on three benchmark datasets. Note that we employed the T5-large and Flan-T5 large models in the Huggingface's transformer library \footnote{\url{https://huggingface.co/docs/transformers/model_doc/t5}} and only the best results obtained during the validation experiments for each model are presented. Note that the experiments were conducted using Nvidia RTX A6000 GPUs. 

\noindent {\bf PUBHEALTH Results:} The PUBHEALTH \cite{kotonya2020explainable} dataset consists of health-related claims with justifications written by journalists which were considered as gold explanations to evaluate the correctness of claims. Each claim was annotated as \textit{True}, \textit{False}, \textit{Mixture} or \textit{Unproven}. The training set consists of 9466 claims and 1183 claims exist in validation and test sets.

In the experiments, the dropout and the learning rate were set to 0.1 and 1e-4 respectively with an Adam optimizer with a linear decay scheduler was employed. In addition, the hidden layer size (denoted as "m" in Figure \ref{fig:t5_mt}) was determined as 128. The weights were assigned as follows: $w_{summary}$ and $w_{classification}$ were set to 0.5, $w_{mixture}$ to 2.5, and $w_{unproven}$ to 7.


\begin{table}[t!]
\centering
\caption{Summarization results on PUBHEALTH}
\label{tab:pubhealth-summary}
\scalebox{0.72}{%
\begin{tabular}{l|lll}
Model          & Rouge-1 & Rouge-2 & Rouge-L \\
\hline
\begin{tabular}[c]{@{}l@{}}Oracle\\ \cite{kotonya2020explainable} \end{tabular}& 39.24   & 14.89   & 32.78   \\
\begin{tabular}[c]{@{}l@{}}Lead-3\\ \cite{kotonya2020explainable} \end{tabular}       & 29.01   & 10.24   & 24.18   \\
\hline
\begin{tabular}[c]{@{}l@{}}EXPLAINERFC-EXPERT\\ \cite{kotonya2020explainable} \end{tabular}
 & 32.30   & 13.46   & 26.99  \\
\hline
T5 single-task  & 30.90   & 13.40   & 27.16 \\
T5 multi-task  & 32.55	  & 14.54   & 28.60 \\
Flan-T5 single-task & \textbf{33.50} & \textbf{14.64} & \textbf{29.54} \\
Flan-T5 multi-task & 32.38 & 14.03 & 28.41
\end{tabular}%
}
\end{table}

To assess the summarization performance of the multi-task model, the ROUGE-N \cite{lin2003automatic} and ROUGE-L \cite{lin2004automatic} metrics were employed. These metrics were utilized to compare the proposed model against a baseline, an oracle, and summarization models implemented by \cite{kotonya2020explainable}. ROUGE-N measures the overlap of n-gram sequences between the ground truth and the given model's output. Likewise, the ROUGE-L metric captures the longest common co-occurring n-gram sequences. 

Table \ref{tab:pubhealth-summary} displays the summarization outcomes of the proposed models in comparison to the baseline and Oracle models. Lead-3 \cite{kotonya2020explainable} served as the baseline that utilized the first three sentences as a summary. Oracle \cite{kotonya2020explainable} was an extractive summary model that served as an upper bound. Additionally, EXPLAINERFC-EXPERT \cite{kotonya2020explainable} was a state-of-the-art single-task abstractive summary generator model which performed slightly better than T5 single-task model. On the other hand, the T5 multi-task model outperformed the state-of-the-art model in all Rouge metrics which implies that multitasking improved the summarization results for tT5 model. Note that the T5 single-task and the T5 multi-task models were almost identical to the model architecture given in Figure \ref{fig:t5_mt} but the classification head of the T5 single-task model was set to 0.

Furthermore, the Flan-T5 multi-task model represents an instruction fine-tuned variant of T5 that performed slightly less effectively than the single-task Flan-T5 (for summarization), but both models outperformed the state-of-the-art model. 


\begin{table}[t!]
\centering
\caption{Veracity results on PUBHEALTH}
\label{tab:pubhealth-classification}
\scalebox{0.62}{%
\begin{tabular}{l|llll}
Model                  & Precision      & Recall         & F1-macro       & Accuracy       \\
\hline
\begin{tabular}[c]{@{}l@{}}BERT (rand. sentences)\\ \cite{kotonya2020explainable} \end{tabular}& 38.97          & 39.38          & 39.16          & 20.99          \\
\begin{tabular}[c]{@{}l@{}}BERT (all sentences)\\ \cite{kotonya2020explainable} \end{tabular} & 56.50          & 56.50          & 56.50          & 56.40          \\
\hline
\begin{tabular}[c]{@{}l@{}}BERT (top-k) \\ \cite{kotonya2020explainable} \end{tabular}           & 77.39 & 54.77          & 63.93          & 66.02          \\
\begin{tabular}[c]{@{}l@{}}SCIBERT\\ \cite{kotonya2020explainable} \end{tabular}             & 75.69          & 66.20          & \textbf{70.52} & 69.73 \\
\hline
T5 single-task  & \textbf{78.24}   & 71.05   & 61.08  & 71.05  \\

T5 multi-task   & 77.62   & 70.32   & 60.93 & 70.32 \\
Flan-T5 single-task & 74.80 & 73.56 & 61.39 & 73.56 \\
Flan-T5 multi-task & 76.46 & \textbf{76.64} & 65.18 & \textbf{76.64}

\end{tabular}%
}
\end{table}

The results for veracity prediction using the precision, recall, F1-macro and accuracy metrics were presented in Table \ref{tab:pubhealth-classification}. The first two rows indicated the baselines. BERT (top-k) and SCIBERT models applied a sentence selection based on the sentences' semantic similarity with the claim sentences. For evidence selection, the authors employed the S-BERT \cite{reimers2019sentence} model. Therefore, we followed a similar approach and selected the top-5 evidence sentences and the claim statement as input for these models.

The results indicate that the Flan-T5 variant outperformed the T5-based models for veracity prediction but on the F1-macro metric the state-of-the-art SCIBERT model performed significantly better than the proposed models. The main reason for this difference can be attributed to the considerable imbalance in label distribution. For instance, the ratio of claims labeled as \textit{Unproven} is approximately 3.2\%, while the \textit{Mixture} cases constitute around 15.2\% of the dataset. Our post-evaluation analysis, in Appendix \ref{sec:cm-appendix}, revealed that despite the usage of additional coefficients for the \textit{Unproven} and \textit{Mixture} instances, our models suffered from the data imbalance problem. Overall, multi-tasking has a positive influence on the performance of our model in the veracity prediction task. Nevertheless, the quality of the abstractive summaries is subject to both positive and negative effects, depending on the selection of the base model.

\noindent {\bf FEVER Results:} 
FEVER \cite{thorne2018fever} is a benchmark dataset that includes 185K claims with related evidence documents from Wikipedia. The dataset was published for the FEVER shared tasks in 2018. For the fact-checking task, the claim statements were annotated as \textit{Supports}, \textit{Refutes} and \textit{Not enough info}. 

Since the FEVER test set did not contain the true labels, the multi-task model's veracity prediction performance was evaluated using the development set. To retrieve evidence documents, DOMLIN system \cite{stammbach2019team} was employed. DOMLIN \cite{stammbach2019team} is a two-stage evidence retrieval system designed to enhance evidence recall. First, the document retrieval module selects sentences that can be considered as potential evidence. Secondly, hyperlinks and the content of the hyperlinked pages are examined to uncover additional evidence. Additionally, the authors utilized BERT-base \cite{devlin2018bert} for evidence retrieval which was upgraded to ROBERTA-base \cite{liu2019roberta} in the enhanced DOMLIN++ \cite{stammbach2020fever} version. 

DOMLIN retrieved evidence documents for 17K out of the 20K claims in the development set, while labeling the remaining instances as "not enough info." With this supporting information, our multi-task model achieved an accuracy score of 76.18\%. However, its Flan-T5-based counterpart outperformed it with a score of 80.44\%. It's worth noting that the DOMLIN model \cite{stammbach2019team} achieved an accuracy of 71.44\%, DOMLIN++ \cite{stammbach2020fever} achieved 77.48\%, and the E-BART \cite{brand2022neural} model reached an accuracy of 75.10\% by utilizing the similar evidence retrieval method.

\begin{table}[t!]
\centering
\caption{Veracity and summarization results on e-FEVER}
\label{tab:efever-results}
\scalebox{0.51}{%
\begin{tabular}{cc|cc|ccc}
\hline
\textbf{Model}                                                                  & \textbf{Dataset}                                             & \textbf{\begin{tabular}[c]{@{}c@{}}Acc.\\ (w/o N.E.I)\end{tabular}} & \textbf{Acc.} & \textbf{Rouge-1} & \textbf{Rouge-2} & \textbf{Rouge-L} \\ \hline
 
\begin{tabular}[c]{@{}l@{}}E-BARTSmall\\ \cite{brand2022neural} \end{tabular}&  eFever\_Small & 87.2  & \textbf{78.2}     & 73.58           & \textbf{64.37}  & 71.43           \\
T5-Small                                                                       & eFever\_Small                                                & \textbf{91.11}                                                                       & 74.75             & 74.00            & 63.64            & 72.78            \\
\begin{tabular}[c]{@{}c@{}}T5-Small\\ (uncertainty \\  weighting)\end{tabular} & eFever\_Small                                                & 90.66                                                                       & 74.57             & \textbf{74.46}   & 64.32            & \textbf{73.19}   \\ 
\hline

\begin{tabular}[c]{@{}c@{}}T5-Full ( Only \\  Summarization)\end{tabular}      & eFever\_Full                                                 & -                                                                           & -                 & 65.94            & 57.53            & 65.09            \\
\begin{tabular}[c]{@{}c@{}}Flan-T5-Full ( Only \\  Summarization)\end{tabular}      & eFever\_Full                                                 & -                                                                           & -                 & 68.79            & 60.87            & 67.92            \\
\begin{tabular}[c]{@{}c@{}}T5-Full (Only \\  Classification)\end{tabular}      & eFever\_Full                                                 & 91.12                                                              & 73.61             & -                & -                & -  \\
\begin{tabular}[c]{@{}c@{}}Flan-T5-Full (Only \\  Classification)\end{tabular}      & eFever\_Full                                                 & 93.94                                                              & 78.87             & -                & -                & -  \\\hline
\begin{tabular}[c]{@{}l@{}}E-BARTFull\\ \cite{brand2022neural} \end{tabular}   & eFever\_Full  & 85.2   & 77.2              & 65.51           & 57.60           & 64.07           \\ 
T5-Full                                                                        & eFever\_Full                                                 & 90.91                                                                       & 75,26             & 68,16            & 59,96            & 67,26            \\
\begin{tabular}[c]{@{}c@{}}T5-Full\\ (uncertainty\\  weighting)\end{tabular}   & eFever\_Full                                                 & 90.90                                                                       & 74,28             & 67,30            & 59,36            & 66,49            \\ 
Flan-T5 & eFever\_Full & \textbf{94.36} & \textbf{79.91} & 66.75 & 58.42 & 65.88 \\
\begin{tabular}[c]{@{}c@{}}Flan-T5\\ (uncertainty\\  weighting)\end{tabular}  & eFever\_Full & 93.94 & 79.02 & \textbf{68.84} & \textbf{60.89} & \textbf{67.97}
\end{tabular}%
}
\end{table}

\noindent {\bf e-FEVER Results:} The e-FEVER dataset \cite{stammbach2020fever} is a subset of the original FEVER dataset and consists of 67687 claims with evidence documents retrieved using the DOMLIN system \cite{stammbach2019team}. In addition to claims and evidence documents, the authors published the summaries using the GPT-3 model \cite{brown2020language} for each claim. Hence, these summaries were leveraged as ground-truth explanations to compare our model's decision-making process with the GPT-3-based model.

The authors pointed out that the GPT-3-based model generated null summaries for certain claims. To address this issue, similar to Brand et al. \cite{brand2022neural}, two variations of the dataset were utilized: \textit{e-FEVER\textunderscore Full} and \textit{e-FEVER\textunderscore Small}. The former contains all claims, while the latter excluded instances with null summaries. The \textit{e-FEVER\textunderscore Small} consists of 40702 instances. Moreover, \cite{brand2022neural} provided some examples labeled as \textit{Not enough info} that could be either refuted or supported based on the provided evidence documents. Therefore, the binary veracity prediction performance of the multi-task model was measured by ignoring the \textit{Not enough info} instances. Likewise, similar to \cite{brand2022neural} two variations of the multi-task model were trained: \textit{T5-Small} and \textit{T5-Full} where the former was trained on \textit{e-FEVER\textunderscore Small} and the latter was trained on \textit{e-FEVER\textunderscore Full}. 

After conducting several validation experiments, the best results were obtained on the e-FEVER dataset by setting the batch size
to 4 and the hidden layer size (denoted as "m" in Figure \ref{fig:t5_mt}) to 32. Moreover, we conducted experiments by employing both static loss weighting and uncertainty loss weighting strategies. For the static loss strategy, the weights were assigned as follows: $w_{summary}$ is set to 0.2 and $w_{classification}$ to 0.8.

Table \ref{tab:efever-results} demonstrated the summarization and veracity prediction results on the e-FEVER dataset. To the best of our knowledge, only \cite{brand2022neural} reported results on this dataset. The first three rows indicated the models that utilized \textit{e-FEVER\textunderscore Small} dataset. Both of the T5-based multi-task models performed slightly better than the E-BARTSmall model for summarization (except Rouge-2) and binary classification. However, E-BARTSmall achieved significantly higher accuracy (78.2\%) than the proposed models in three-class classification. 

Secondly, the baseline models were outlined starting from the fourth row to the seventh row which were trained specifically for either summarization or classification. Therefore, we did not report the classification results for the summarization model, and vice versa. Similarly, on \textit{eFever\textunderscore Full} the multi-task T5 models achieved higher binary classification accuracy and summarization scores but performed worse than the E-BARTFull model (77.2\%) in multi-class classification. On the other hand, replacing T5 with the Flan-T5 version led to the highest accuracy scores in both binary and multi-class classification (94.36\% and 79.91\% respectively). Moreover, the Rouge scores of the T5 and Flan-T5 models were higher than the E-BART model on the \textit{eFever\textunderscore Full} dataset. 

Furthermore, we also evaluated the impact of the loss strategy. To be more specific, we employed static loss weighting and uncertainty loss weighting which dynamically adapts the loss weights during training. According to the results, with uncertainty loss weighting the multi-task models performed slightly better on summarization but performed slightly worse on classification on both \textit{eFever\textunderscore Small} and \textit{eFever\textunderscore Full}. Overall, similar to the PUBHEALTH results, the multi-task models based on Flan-T5 demonstrated improved performance in classification through joint training. However, there was a slight decline in summary quality with multi-tasking. Conversely, T5-based models significantly improved on summarization with the aid of multi-tasking but decreased slightly in binary prediction accuracy.

%% file: sections/5-conclusion.tex
\section{Conclusion}

In this paper, we assess the effectiveness of multi-task training of veracity prediction and text summarization. More specifically, we formulated text summarization as the explanation of the veracity prediction task and introduced a T5-based explainable multi-task fact-checking model. The experimental results reveal that in the case of the Flan-T5 model, joint training for summarization and text classification enhances performance in text classification, but with a slight decrease in summarization results. Conversely, for the T5 model, there is a notable increase in summarization results, while the impact on classification results is not as substantial.

In future work, we aim to conduct a user study to evaluate the model's explanations' coherence and quality and also assess the explanations with the related studies. Moreover, transformer-based language models demand substantial computational and hardware resources. However, some recent parameter-efficient fine-tuning techniques, such as LoRA \cite{hu2022lora}, have demonstrated their effectiveness. We also aim to adapt such techniques to enhance overall applicability.

%% file: appendix.tex
\section{Confusion Matrices}
\label{sec:cm-appendix}

\begin{table}[h!]
\centering
\caption{Confusion Matrix}
\label{tab:t5_conf}
\scalebox{0.65}{%
\begin{tabular}{l|l|llll|l}
Model     &    & Unproven & False & Mixture & True & Accuracy \\ \hline
T5 & Unproven & 27       & 8    & 5       & 5  & 60.00 \\ 
single & False    & 31       & 244   & 94     & 19  & 62.89 \\ 
task & Mixture  & 17       & 41    & 131    & 12  & 65.17 \\
&True     & 21       & 8     & 96     & 474 & 79.13 \\ 
\hline
T5 & Unproven & 26       & 10    & 4       & 5   & 57.78\\ 
multi & False    & 31       & 236   & 106      & 15  & 62.43 \\ 
task & Mixture  & 13       & 37    & 137      & 14  & 68.16 \\
&True     & 15       & 17    & 99      & 468 & 78.13 \\ 
\hline
Flan-T5 & Unproven & 25       & 14    & 1       & 5   & 55.56 \\ 
multi & False    & 14       & 307   & 48      & 19  & 79.12 \\ 
task & Mixture  & 9       & 61    & 87      & 44  & 43.28 \\
&True     & 9       & 25    & 39      & 526 & 87.81 \\ 
\end{tabular}%
}
\end{table}

The confusion matrices of the models given in Table \ref{tab:pubhealth-classification} are demonstrated in Table \ref{tab:t5_conf}. Confusion matrices revealed that the margins between the state-of-the-art model's and our models' F1-macro scores are attributed to the class distributions. More specifically, the dataset is highly imbalanced and despite boosting the \textit{Unproven} and \textit{Mixture} instances, the models suffered from the class imbalance problem. Moreover, another takeaway is that boosting the \textit{Mixture} instances decreased the accuracy of \textit{False} claims, particularly for T5 models.

\begin{table*}[t!]
\centering
\caption{Grid search of loss coefficients}
\label{tab:sweep_results}
\scalebox{0.75}{%
\begin{tabular}{cc|ccc|cc}
\textbf{\begin{tabular}[c]{@{}l@{}}Veracity (a), \\ Summary (b) \\ loss coefficients\end{tabular}} & \textbf{Veracity label coefficients}       & \textbf{Rouge-1} & \textbf{Rouge-2} & \textbf{Rouge-L} & \textbf{F1-macro} & \textbf{F1-weighted} \\ \hline
a=0.7, b=0.3                                                                                       &  mixture\_coeff=1.75, unproven\_coeff=5 & 31,99            & 14,14            & 28,18            & 51,14             & 66,66                \\
a=0.7, b=0.3                                                                                       & mixture\_coeff=1.75, unproven\_coeff=7 & 31,93            & 14,26            & 28,46            & \textbf{60,76}    & \textbf{73,16}       \\
a=0.7, b=0.3                                                                                       & mixture\_coeff=1.75, unproven\_coeff=9 & 31,87            & 14,16            & 28,13            & 48,62             & 63,93                \\
a=0.6, b=0.4                                                                                        & mixture\_coeff=1.75, unproven\_coeff=5 & 31,25            & 13,81            & 27,59            & 54,71             & 69,92                \\
a=0.6, b=0.4                                                                                        &  mixture\_coeff=1.75, unproven\_coeff=7 & 32,36            & \textbf{14,59}   & 28,67            & 57,22             & 71,47                \\
a=0.6, b=0.4                                                                                        & mixture\_coeff=1.75, unproven\_coeff=9 & 31,85            & 14,21            & 28,16            & 54,14             & 68,48                \\
a=0.5, b=0.5                                                                                        &  mixture\_coeff=1.75, unproven\_coeff=5 & 32,52            & 14,50            & \textbf{28,74}   & 56,71             & 69,86                \\
a=0.5, b=0.5                                                                                        & mixture\_coeff=1.75, unproven\_coeff=7 & 31,87            & 13,94            & 27,09            & 52,00             & 67,25                \\
a=0.5, b=0.5                                                                                        & mixture\_coeff=1.75, unproven\_coeff=9 & 31,71            & 13,88            & 28,19            & 51,12             & 65,78                \\
a=0.7, b=0.3                                                                                       &  mixture\_coeff=2.5, unproven\_coeff=5  & 31,02            & 13,50            & 27,53            & 50,94             & 65,48                \\
a=0.7, b=0.3                                                                                       &  mixture\_coeff=2.5, unproven\_coeff=7  & 31,82            & 14,00            & 28,12            & 55,87             & 68,57                \\
a=0.7, b=0.3                                                                                       & mixture\_coeff=2.5, unproven\_coeff=9 & 31,96            & 14,42            & 28,40            & 56,52             & 69,93                \\
a=0.6, b=0.4                                                                                        & mixture\_coeff=2.5, unproven\_coeff=5  & 31,43            & 14,03            & 27,75            & 50,59             & 65,76                \\
a=0.6, b=0.4                                                                                        &  mixture\_coeff=2.5, unproven\_coeff=7  & 31,96            & 14,38            & 28,28            & 55,62             & 68,57                \\
a=0.6, b=0.4                                                                                        &  mixture\_coeff=2.5, unproven\_coeff=9  & \textbf{32,54}   & 14,48            & 28,69            & 60,07             & 72,50                \\
a=0.5, b=0.5                                                                                        & mixture\_coeff=2.5, unproven\_coeff=5  & 31,78            & 13,86            & 28,04            & 58,73             & 72,20                \\
a=0.8, b=0.2                                                                                        & mixture\_coeff=1.75, unproven\_coeff=5 & 32,27            & 14,32            & 28,64            & 58,02             & 72,19                \\
a=0.8, b=0.2                                                                                        &  mixture\_coeff=1.75, unproven\_coeff=7 & 31,05            & 13,44            & 27,49            & 50,96             & 65,48                \\
a=0.8, b=0.2                                                                                        &  mixture\_coeff=1.75, unproven\_coeff=9 & 32,03            & 13,74            & 28,06            & 57,59             & 70,41                \\
a=0.5, b=0.5                                                                                       &  mixture\_coeff=1.75, unproven\_coeff=5 & 32,00            & 14,29            & 28,38            & 56,31             & 70,19                \\
a=0.5, b=0.5                                                                                       &  mixture\_coeff=1.75, unproven\_coeff=7 & 31,82            & 14,16            & 28,14            & 55,05             & 69,52                \\
a=0.5, b=0.5                                                                                       &  mixture\_coeff=1.75, unproven\_coeff=9 & 31,87            & 14,15            & 28,22            & 58,33             & 72,34                \\
a=0.8, b=0.2                                                                                        &  mixture\_coeff=2.5, unproven\_coeff=5  & 32,42            & 14,11            & 28,50            & 54,14             & 67,34                \\
a=0.8, b=0.2                                                                                        & mixture\_coeff=2.5, unproven\_coeff=7  & 32,03            & 14,20            & 28,31            & 58,87             & 71,89                \\
a=0.8, b=0.2                                                                                        &  mixture\_coeff=2.5, unproven\_coeff=9  & 31,84            & 13,93            & 28,07            & 58,10             & 71,95                \\
a=0.5, b=0.5                                                                                       &  mixture\_coeff=2.5, unproven\_coeff=5  & 31,85            & 14,25            & 28,13            & 52,58             & 66,45                \\
a=0.5, b=0.5                                                                                       &  mixture\_coeff=2.5, unproven\_coeff=7                       & 32,33            & 14,18            & 28,48            & 60,33             & 73,11                \\
a=0.5, b=0.5                                                                                       & mixture\_coeff=2.5, unproven\_coeff=9  & 31,90            & 14,14            & 28,27            & 55,56             & 70,32                
\end{tabular}%
}
\end{table*}

\section{Grid Search of Static Loss Coefficients}
\label{sec:slc-appendix}


We performed an ablation study to explore candidate values to find an optimal set of hyper-parameters for our multi-task model. We performed a grid search using PUBHEALTH \cite{kotonya2020explainable} dataset to determine the optimal set of loss coefficients. The experimental results are presented in Table \ref{tab:sweep_results}. Note that, we kept the linear layers' size (for veracity prediction), dropout probability, batch size and number of epoch constant.

\section{Grid Search of Hidden Layer Dimensions for Veracity Prediction}
\label{sec:hls-appendix}

We also performed another ablation study to discover the optimal hidden layer size of the classification head of our multi-task model using the PUBHEALTH \cite{kotonya2020explainable} dataset. The experimental results are presented in Table \ref{tab:sweep_hidden}. Note that, we kept the dropout probability, batch size and number of epochs constant.

\begin{table*}[t!]
\centering
\caption{Grid search of hidden layer size}
\label{tab:sweep_hidden}
\scalebox{0.75}{%
\begin{tabular}{@{}ccc|ccc|cc@{}}
\textbf{\begin{tabular}[c]{@{}l@{}}Veracity (a), \\ Summary (b)\\ loss coefficients\end{tabular}} & \textbf{Veracity label coefficients}                                                 & \textbf{\begin{tabular}[c]{@{}l@{}}Hidden \\ Dim\end{tabular}} & \textbf{Rouge-1} & \textbf{Rouge-2} & \textbf{Rouge-L} & \textbf{F1-macro} & \textbf{F1-weighted} \\
\hline
a=0.7, b=0.3                                                                                    &  mixture\_coeff=2.5, unproven\_coeff=7 & 16                  & 31,82            & 14,00            & 28,12            & 55,87             & 68,57                \\
a=0.7, b=0.3                                                                                    &  mixture\_coeff=2.5, unproven\_coeff=9 & 16                  & 31,96            & 14,42            & 28,40            & 56,52             & 69,93                \\
a=0.6, b=0.4                                                                                    &  mixture\_coeff=2.5, unproven\_coeff=7 & 16                  & 31,96            & 14,38            & 28,28            & 55,62             & 68,57                \\
a=0.6, b=0.4                                                                                    &  mixture\_coeff=2.5, unproven\_coeff=9 & 16                  & 32,54            & 14,48            & 28,69            & 60,07             & 72,50                \\
a=0.5, b=0.5                                                                                    &  mixture\_coeff=2.5, unproven\_coeff=7 & 16                  & 32,33            & 14,18            & 28,48            & 60,33             & \textbf{73,11}       \\
a=0.5, b=0.5                                                                                    &  mixture\_coeff=2.5, unproven\_coeff=9 & 16                  & 31,90            & 14,14            & 28,27            & 55,56             & 70,32                \\
a=0.7, b=0.3                                                                                    &  mixture\_coeff=2.5, unproven\_coeff=7 & 32                  & 31,97            & 14,21            & 28,23            & 51,14             & 65,77                \\
a=0.7, b=0.3                                                                                    &  mixture\_coeff=2.5, unproven\_coeff=9 & 32                  & 31,83            & 14,00            & 28,05            & 57,25             & 68,34                \\
a=0.6, b=0.4                                                                                    &  mixture\_coeff=2.5, unproven\_coeff=7 & 32                  & 31,82            & 14,21            & 28,14            & 58,96             & 60,78                \\
a=0.6, b=0.4                                                                                    &  mixture\_coeff=2.5, unproven\_coeff=9 & 32                  & 32,08            & 14,09            & 28,34            & 52,47             & 65,67                \\
a=0.5, b=0.5                                                                                    &  mixture\_coeff=2.5, unproven\_coeff=7 & 32                  & 32,07            & 14,33            & 28,32            & 59,18             & 71,91                \\
a=0.5, b=0.5                                                                                    &  mixture\_coeff=2.5, unproven\_coeff=9 & 32                  & 31,79            & 14,13            & 28,29            & 49,99             & 61,82                \\
a=0.7, b=0.3                                                                                    &  mixture\_coeff=2.5, unproven\_coeff=7 & 64                  & 32,55            & 14,54            & 28,60            & \textbf{60,93}    & 72,51                \\
a=0.7, b=0.3                                                                                    & mixture\_coeff=2.5, unproven\_coeff=9 & 64                  & \textbf{32,69}   & \textbf{14,71}   & \textbf{28,84}   & 49,08             & 62,63                \\
a=0.6, b=0.4                                                                                    &  mixture\_coeff=2.5, unproven\_coeff=7 & 64                  & 31,97            & 14,28            & 28,30            & 44,73             & 57,52                \\
a=0.6, b=0.4                                                                                    &  mixture\_coeff=2.5, unproven\_coeff=9 & 64                  & 31,98            & 14,19            & 28,33            & 57,78             & 72,52                \\
a=0.5, b=0.5                                                                                    &  mixture\_coeff=2.5, unproven\_coeff=7 & 64                  & 31,78            & 13,95            & 28,01            & 59,22             & 72,20                \\
a=0.5, b=0.5                                                                                    &  mixture\_coeff=2.5, unproven\_coeff=9 & 64                  & 31,63            & 13,99            & 27,89            & 53,21             & 66,03                \\ 
a=0.7, b=0.3                                                                                      &  mixture\_coeff=2.5, unproven\_coeff=7 & 128                                                            & 31,97            & 14,21            & 28,23            & 51,14             & 65,77                \\
a=0.7, b=0.3                                                                                      & mixture\_coeff=2.5, unproven\_coeff=9 & 128                                                            & 31,83            & 14,00            & 28,05            & 57,25             & 68,34                \\
a=0.6, b=0.4                                                                                      &  mixture\_coeff=2.5, unproven\_coeff=7 & 128                                                            & 31,82            & 14,21            & 28,14            & 48,42             & 60,78                \\
a=0.6, b=0.4                                                                                      &  mixture\_coeff=2.5, unproven\_coeff=9 & 128                                                            & 32,08            & 14,09            & 28,34            & 52,47             & 65,67                \\
a=0.5, b=0.5                                                                                      &  mixture\_coeff=2.5, unproven\_coeff=7 & 128                                                            & 31,79            & 14,13            & 28,29            & 49,99             & 61,82                \\
a=0.5, b=0.5                                                                                      &  mixture\_coeff=2.5, unproven\_coeff=9 & 128                                                            & 32,55            & 14,54            & 28,60            & \textbf{60,93}    & 72,51                \\
a=0.7, b=0.3                                                                                      &  mixture\_coeff=2.5, unproven\_coeff=7 & 256                                                            & 32,07            & 14,33            & 28,32            & 59,18             & 71,91                \\
a=0.7, b=0.3                                                                                      &  mixture\_coeff=2.5, unproven\_coeff=9 & 256                                                            & \textbf{32,69}   & \textbf{14,71}   & \textbf{28,84}   & 49,08             & 62,63                \\
a=0.6, b=0.4                                                                                      & mixture\_coeff=2.5, unproven\_coeff=7 & 256                                                            & 31,97            & 14,28            & 28,30            & 44,73             & 57,52               

\end{tabular}%
}
\end{table*}

\section{More Examples for Generated Summaries}
\label{sec:summary-appendix}
\noindent\textbf{EXAMPLE 1} \\
\noindent\textbf{Claim:}  “While California is dying … Gavin (Newsom) is vacationing in Stevensville, MT!” \\
\noindent\textbf{Evidence:}  ... A Facebook post said, "While California is dying … Gavin (Newsom) is vacationing in Stevensville, MT!" There is no evidence of this. Newsom’s office said he has not been vacationing in Stevensville, and so did the lieutenant governor’s office, the Montana governor’s office and the Stevensville mayor. Newsom gave a live press conference from a California restaurant on May 18, the day the post went up.\\
\noindent\textbf{Gold Summary:} Newsom has not been vacationing in Stevensville in recent weeks, his office said. The Montana governor’s office and Stevensville mayor said the same thing. \\
\noindent\textbf{Generated Summary:}  A Facebook post said Newsom was vacationing in Stevensville, Montana. The governor has not been there in recent weeks.\\
\noindent\textbf{Gold Label:} FALSE\\
\noindent\textbf{Predicted Label:}  FALSE \\

\noindent\textbf{EXAMPLE 2} \\
\noindent\textbf{Claim:}   Treating at the Earliest Sign of MS May Offer Long-Term Benefit \\
\noindent\textbf{Evidence:}  ... The study was addressing whether early therapy was better than later therapy, so the release should not be held to the standard of comparing interferon to other drugs. However, it’s worth noting here that the number of drugs for treating MS have nearly doubled to 20 over the past three years. The drug is widely available and the news release gives that impression as well. The release explains that this particular study is adding to evidence and is not particularly novel. “Our study adds to the evidence supporting treatment at the earliest sign of the disease and indicates that early treatment has a long-lasting effect on disease activity,” a researcher says in the release. However, another novelty of the study is starting treatment after one event and not waiting for any further events. This could lead to a much larger market for the drug. It could be concerning since some patients only have a few recurrences in their lifetime. There was no unjustifiable language. \\
\noindent\textbf{Gold Summary:} This release summarizes a study that followed 11-year treatment outcomes of two groups of patients with early symptoms of multiple sclerosis (MS). All of the patients had experienced a first episode suggestive of MS, called clinically isolated syndrome (CIS), and a  magnetic resonance imaging (MRI) scan suggesting MS. Up to 85 percent of people with CIS go on to develop MS. The two patient groups were randomized to either receive interferon beta-1 by injection immediately or receive a placebo for two years. After two years, the placebo group could receive their treatment of choice. The patients who received the drug immediately appeared to have fewer relapses over the 11 years of study, according to the lead author, but no advantages were observed in other patient outcomes such as overall disability and MRI scans. The release would have been stronger if it provided more context for judging the reduced number of relapses, especially in light of the two groups showing no difference in disability. The release also omitted mention of significant potential conflicts of interest, the high annual cost of interferon beta-1 and the drug’s side effects. Society faces tough questions because of the high prices of drugs that target the immune system, such as interferon beta-1b for multiple sclerosis. This news release suggests drug therapy should begin sooner for patients without including any of the controversy on cost playing out in the United States and Europe. Last year authors of a different research study, also published in Neurology, called the rise in MS drug prices “alarming” and wrote it was also puzzling, since some of the drugs are not new. “Why the costs of MS DMTs [disease modifying therapies] in the United States have risen so dramatically is uncertain. However, the simplest explanation is that pharmaceutical companies raise prices of new and old MS DMTs in the United States to increase profits and our health care system puts no limits on these increases,’ they wrote. Why is cost so important to mention? This drug study, in particular, has significant potential conflicts of interest, discussed below under the Funding Sources \& Conflict of Interest Disclosure criteria. \\
\noindent\textbf{Generated Summary:}  This news release describes a study that found early treatment with interferon beta-1b (brand name Betaseron) for patients with demyelination, often an early sign of multiple sclerosis (MS), was associated with a lower rate of relapses at 11 years. The release does a good job of explaining the study and its limitations. However, it did not mention any potential harms of treatment. There are significant side effects reported to therapy with interferon beta-1b, including depression and thoughts of suicide, according to the National Library of Medicine. No harms were mentioned in the study. We also wonder — if patients routinely receive a therapy before official diagnosis — whether that risks treating people for a condition they do not eventually develop. The published study states that about 85\% of people who encounter a single episode of demyelination (often an early sign of MS) go on to develop MS. \\
\noindent\textbf{Gold Label:} MIXTURE\\
\noindent\textbf{Predicted Label:}  MIXTURE

\noindent\textbf{EXAMPLE 3} \\
\noindent\textbf{Claim:}  Queen Elizabeth II wore a Burmese Ruby Tiara as a slight against U.S. President Donald Trump. \\
\noindent\textbf{Evidence:}  ...  Given that Queen Elizabeth has worn this tiara in the company of a variety of world leaders and for a number of different special occasions, those claiming that she chose to wear it in June 2019 specifically as a slight against President Trump appear to be revealing more about their own political beliefs than the queen’s.\\
\noindent\textbf{Gold Summary:} What's true: Queen Elizabeth II wore a Burmese Ruby Tiara while meeting President Donald Trump in June 2019. The 96 rubies that adorn this tiara are said to symbolically protect the wearer from 96 diseases. What's false: No evidence exists that the queen specifically chose this tiara as a slight against Trump, and the queen has worn this same tiara on several other occasions and in the company of a wide range of world leaders. \\
\noindent\textbf{Generated Summary:}  Did Queen Elizabeth II Wear a Burmese Ruby Tiara as a Sighting of Disrespect? \\
\noindent\textbf{Gold Label:} FALSE\\
\noindent\textbf{Predicted Label:}  UNPROVEN \\

%% file: acl_latex.bbl
\begin{thebibliography}{56}
\expandafter\ifx\csname natexlab\endcsname\relax\def\natexlab#1{#1}\fi

\bibitem[{Ahmadi et~al.(2020)Ahmadi, Truong, Dao, Ortona, and Papotti}]{ahmadi2020rulehub}
Naser Ahmadi, Thi-Thuy-Duyen Truong, Le-Hong-Mai Dao, Stefano Ortona, and Paolo Papotti. 2020.
\newblock Rulehub: A public corpus of rules for knowledge graphs.
\newblock \emph{Journal of Data and Information Quality (JDIQ)}, 12(4):1--22.

\bibitem[{Allcott and Gentzkow(2017)}]{allcott2017social}
Hunt Allcott and Matthew Gentzkow. 2017.
\newblock Social media and fake news in the 2016 election.
\newblock \emph{Journal of economic perspectives}, 31(2):211--36.

\bibitem[{Atanasova et~al.(2020{\natexlab{a}})Atanasova, Simonsen, Lioma, and Augenstein}]{atanasova2020generating}
Pepa Atanasova, Jakob~Grue Simonsen, Christina Lioma, and Isabelle Augenstein. 2020{\natexlab{a}}.
\newblock Generating fact checking explanations.
\newblock In \emph{Proceedings of the 58th Annual Meeting of the Association for Computational Linguistics}, pages 7352--7364.

\bibitem[{Atanasova et~al.(2020{\natexlab{b}})Atanasova, Wright, and Augenstein}]{atanasova2020adversarial}
Pepa Atanasova, Dustin Wright, and Isabelle Augenstein. 2020{\natexlab{b}}.
\newblock Generating label cohesive and well-formed adversarial claims.
\newblock In \emph{Proceedings of the 2020 Conference on Empirical Methods in Natural Language Processing (EMNLP)}, pages 3168--3177.

\bibitem[{Brand et~al.(2022)Brand, Roitero, Soprano, Rahimi, and Demartini}]{brand2022neural}
Erik Brand, Kevin Roitero, Michael Soprano, Afshin Rahimi, and Gianluca Demartini. 2022.
\newblock A neural model to jointly predict and explain truthfulness of statements.
\newblock \emph{ACM Journal of Data and Information Quality}, 15(1):1--19.

\bibitem[{Brown et~al.(2020)Brown, Mann, Ryder, Subbiah, Kaplan, Dhariwal, Neelakantan, Shyam, Sastry, Askell et~al.}]{brown2020language}
Tom Brown, Benjamin Mann, Nick Ryder, Melanie Subbiah, Jared~D Kaplan, Prafulla Dhariwal, Arvind Neelakantan, Pranav Shyam, Girish Sastry, Amanda Askell, et~al. 2020.
\newblock Language models are few-shot learners.
\newblock \emph{Advances in neural information processing systems}, 33:1877--1901.

\bibitem[{Chen et~al.(2021)Chen, Zhang, and Yang}]{chen2021multi}
Shijie Chen, Yu~Zhang, and Qiang Yang. 2021.
\newblock Multi-task learning in natural language processing: An overview.
\newblock \emph{arXiv preprint arXiv:2109.09138}.

\bibitem[{Cheng et~al.(2021)Cheng, Guo, Shu, and Liu}]{cheng2021causal}
Lu~Cheng, Ruocheng Guo, Kai Shu, and Huan Liu. 2021.
\newblock Causal understanding of fake news dissemination on social media.
\newblock In \emph{Proceedings of the 27th ACM SIGKDD Conference on Knowledge Discovery \& Data Mining}, pages 148--157.

\bibitem[{Chung et~al.(2022)Chung, Hou, Longpre, Zoph, Tay, Fedus, Li, Wang, Dehghani, Brahma et~al.}]{chung2022scaling}
Hyung~Won Chung, Le~Hou, Shayne Longpre, Barret Zoph, Yi~Tay, William Fedus, Eric Li, Xuezhi Wang, Mostafa Dehghani, Siddhartha Brahma, et~al. 2022.
\newblock Scaling instruction-finetuned language models.
\newblock \emph{arXiv preprint arXiv:2210.11416}.

\bibitem[{Crawshaw(2020)}]{crawshaw2020multi}
Michael Crawshaw. 2020.
\newblock Multi-task learning with deep neural networks: A survey.
\newblock \emph{arXiv preprint arXiv:2009.09796}.

\bibitem[{Cui et~al.(2020)Cui, Seo, Tabar, Ma, Wang, and Lee}]{cui2020deterrent}
Limeng Cui, Haeseung Seo, Maryam Tabar, Fenglong Ma, Suhang Wang, and Dongwon Lee. 2020.
\newblock Deterrent: Knowledge guided graph attention network for detecting healthcare misinformation.
\newblock In \emph{Proceedings of the 26th ACM SIGKDD international conference on knowledge discovery \& data mining}, pages 492--502.

\bibitem[{Dai et~al.(2022)Dai, Hsu, Xiong, and Ku}]{dai2022ask}
Shih-Chieh Dai, Yi-Li Hsu, Aiping Xiong, and Lun-Wei Ku. 2022.
\newblock Ask to know more: Generating counterfactual explanations for fake claims.
\newblock \emph{arXiv preprint arXiv:2206.04869}.

\bibitem[{Devlin et~al.(2018)Devlin, Chang, Lee, and Toutanova}]{devlin2018bert}
Jacob Devlin, Ming-Wei Chang, Kenton Lee, and Kristina Toutanova. 2018.
\newblock Bert: Pre-training of deep bidirectional transformers for language understanding.
\newblock \emph{arXiv preprint arXiv:1810.04805}.

\bibitem[{Gad-Elrab et~al.(2019)Gad-Elrab, Stepanova, Urbani, and Weikum}]{gad2019exfakt}
Mohamed~H Gad-Elrab, Daria Stepanova, Jacopo Urbani, and Gerhard Weikum. 2019.
\newblock Exfakt: A framework for explaining facts over knowledge graphs and text.
\newblock In \emph{Proceedings of the Twelfth ACM International Conference on Web Search and Data Mining}, pages 87--95.

\bibitem[{Guo et~al.(2022)Guo, Schlichtkrull, and Vlachos}]{guo2022survey}
Zhijiang Guo, Michael Schlichtkrull, and Andreas Vlachos. 2022.
\newblock A survey on automated fact-checking.
\newblock \emph{Transactions of the Association for Computational Linguistics}, 10:178--206.

\bibitem[{Hartmann et~al.(2019)Hartmann, Golovchenko, and Augenstein}]{hartmann2019mapping}
Mareike Hartmann, Yevgeniy Golovchenko, and Isabelle Augenstein. 2019.
\newblock Mapping (dis-) information flow about the mh17 plane crash.
\newblock In \emph{Proceedings of the Second Workshop on Natural Language Processing for Internet Freedom: Censorship, Disinformation, and Propaganda}, pages 45--55.

\bibitem[{Hu et~al.(2022)Hu, Shen, Wallis, Allen-Zhu, Li, Wang, Wang, and Chen}]{hu2022lora}
Edward~J Hu, Yelong Shen, Phillip Wallis, Zeyuan Allen-Zhu, Yuanzhi Li, Shean Wang, Lu~Wang, and Weizhu Chen. 2022.
\newblock \href {https://openreview.net/forum?id=nZeVKeeFYf9} {Lo{RA}: Low-rank adaptation of large language models}.
\newblock In \emph{International Conference on Learning Representations}.

\bibitem[{Jin et~al.(2016)Jin, Cao, Zhang, Zhou, and Tian}]{jin2016novel}
Zhiwei Jin, Juan Cao, Yongdong Zhang, Jianshe Zhou, and Qi~Tian. 2016.
\newblock Novel visual and statistical image features for microblogs news verification.
\newblock \emph{IEEE transactions on multimedia}, 19(3):598--608.

\bibitem[{Jwa et~al.(2019)Jwa, Oh, Park, Kang, and Lim}]{jwa2019exbake}
Heejung Jwa, Dongsuk Oh, Kinam Park, Jang~Mook Kang, and Heuiseok Lim. 2019.
\newblock exbake: Automatic fake news detection model based on bidirectional encoder representations from transformers (bert).
\newblock \emph{Applied Sciences}, 9(19):4062.

\bibitem[{Kendall et~al.(2018)Kendall, Gal, and Cipolla}]{kendall2018multi}
Alex Kendall, Yarin Gal, and Roberto Cipolla. 2018.
\newblock Multi-task learning using uncertainty to weigh losses for scene geometry and semantics.
\newblock In \emph{Proceedings of the IEEE conference on computer vision and pattern recognition}, pages 7482--7491.

\bibitem[{Kotonya and Toni(2020{\natexlab{a}})}]{kotonya2020explainablesurvey}
Neema Kotonya and Francesca Toni. 2020{\natexlab{a}}.
\newblock Explainable automated fact-checking: A survey.
\newblock In \emph{Proceedings of the 28th International Conference on Computational Linguistics}, pages 5430--5443.

\bibitem[{Kotonya and Toni(2020{\natexlab{b}})}]{kotonya2020explainable}
Neema Kotonya and Francesca Toni. 2020{\natexlab{b}}.
\newblock Explainable automated fact-checking for public health claims.
\newblock In \emph{Proceedings of the 2020 Conference on Empirical Methods in Natural Language Processing (EMNLP)}, pages 7740--7754.

\bibitem[{Lewis et~al.(2020)Lewis, Liu, Goyal, Ghazvininejad, Mohamed, Levy, Stoyanov, and Zettlemoyer}]{lewis-etal-2020-bart}
Mike Lewis, Yinhan Liu, Naman Goyal, Marjan Ghazvininejad, Abdelrahman Mohamed, Omer Levy, Veselin Stoyanov, and Luke Zettlemoyer. 2020.
\newblock \href {https://doi.org/10.18653/v1/2020.acl-main.703} {{BART}: Denoising sequence-to-sequence pre-training for natural language generation, translation, and comprehension}.
\newblock In \emph{Proceedings of the 58th Annual Meeting of the Association for Computational Linguistics}, pages 7871--7880, Online. Association for Computational Linguistics.

\bibitem[{Li et~al.(2023)Li, Lee, Kordzadeh, and Guo}]{li2023boosts}
Yichuan Li, Kyumin Lee, Nima Kordzadeh, and Ruocheng Guo. 2023.
\newblock What boosts fake news dissemination on social media? a causal inference view.
\newblock In \emph{Pacific-Asia Conference on Knowledge Discovery and Data Mining}, pages 234--246. Springer.

\bibitem[{Lin and Hovy(2003)}]{lin2003automatic}
Chin-Yew Lin and Eduard Hovy. 2003.
\newblock Automatic evaluation of summaries using n-gram co-occurrence statistics.
\newblock In \emph{Proceedings of the 2003 human language technology conference of the North American chapter of the association for computational linguistics}, pages 150--157.

\bibitem[{Lin and Och(2004)}]{lin2004automatic}
Chin-Yew Lin and Franz~Josef Och. 2004.
\newblock Automatic evaluation of machine translation quality using longest common subsequence and skip-bigram statistics.
\newblock In \emph{Proceedings of the 42nd Annual Meeting of the Association for Computational Linguistics (ACL-04)}, pages 605--612.

\bibitem[{Liu et~al.(2019)Liu, Ott, Goyal, Du, Joshi, Chen, Levy, Lewis, Zettlemoyer, and Stoyanov}]{liu2019roberta}
Yinhan Liu, Myle Ott, Naman Goyal, Jingfei Du, Mandar Joshi, Danqi Chen, Omer Levy, Mike Lewis, Luke Zettlemoyer, and Veselin Stoyanov. 2019.
\newblock Roberta: A robustly optimized bert pretraining approach.
\newblock \emph{arXiv preprint arXiv:1907.11692}.

\bibitem[{Lu and Li(2020)}]{lu2020gcan}
Yi-Ju Lu and Cheng-Te Li. 2020.
\newblock Gcan: Graph-aware co-attention networks for explainable fake news detection on social media.
\newblock In \emph{Proceedings of the 58th Annual Meeting of the Association for Computational Linguistics}, pages 505--514.

\bibitem[{Oshikawa et~al.(2020)Oshikawa, Qian, and Wang}]{oshikawa2020survey}
Ray Oshikawa, Jing Qian, and William~Yang Wang. 2020.
\newblock A survey on natural language processing for fake news detection.
\newblock In \emph{Proceedings of the 12th Language Resources and Evaluation Conference}, pages 6086--6093.

\bibitem[{Ousidhoum et~al.(2022)Ousidhoum, Yuan, and Vlachos}]{ousidhoum-etal-2022-varifocal}
Nedjma Ousidhoum, Zhangdie Yuan, and Andreas Vlachos. 2022.
\newblock \href {https://doi.org/10.18653/v1/2022.emnlp-main.163} {Varifocal question generation for fact-checking}.
\newblock In \emph{Proceedings of the 2022 Conference on Empirical Methods in Natural Language Processing}, pages 2532--2544, Abu Dhabi, United Arab Emirates. Association for Computational Linguistics.

\bibitem[{Pan et~al.(2018)Pan, Pavlova, Li, Li, Li, and Liu}]{pan2018content}
Jeff~Z Pan, Siyana Pavlova, Chenxi Li, Ningxi Li, Yangmei Li, and Jinshuo Liu. 2018.
\newblock Content based fake news detection using knowledge graphs.
\newblock In \emph{International semantic web conference}, pages 669--683. Springer.

\bibitem[{Pan et~al.(2023)Pan, Wu, Lu, Luu, Wang, Kan, and Nakov}]{pan-etal-2023-fact}
Liangming Pan, Xiaobao Wu, Xinyuan Lu, Anh~Tuan Luu, William~Yang Wang, Min-Yen Kan, and Preslav Nakov. 2023.
\newblock \href {https://doi.org/10.18653/v1/2023.acl-long.386} {Fact-checking complex claims with program-guided reasoning}.
\newblock In \emph{Proceedings of the 61st Annual Meeting of the Association for Computational Linguistics (Volume 1: Long Papers)}, pages 6981--7004, Toronto, Canada. Association for Computational Linguistics.

\bibitem[{P{\'e}rez-Rosas et~al.(2018)P{\'e}rez-Rosas, Kleinberg, Lefevre, and Mihalcea}]{perez2018automatic}
Ver{\'o}nica P{\'e}rez-Rosas, Bennett Kleinberg, Alexandra Lefevre, and Rada Mihalcea. 2018.
\newblock Automatic detection of fake news.
\newblock In \emph{Proceedings of the 27th International Conference on Computational Linguistics}, pages 3391--3401.

\bibitem[{Pogue(2017)}]{pogue2017stamp}
David Pogue. 2017.
\newblock How to stamp out fake news.
\newblock \emph{Scientific American}, 316(2):24--24.

\bibitem[{Popat et~al.(2017)Popat, Mukherjee, Str{\"o}tgen, and Weikum}]{popat2017truth}
Kashyap Popat, Subhabrata Mukherjee, Jannik Str{\"o}tgen, and Gerhard Weikum. 2017.
\newblock Where the truth lies: Explaining the credibility of emerging claims on the web and social media.
\newblock In \emph{Proceedings of the 26th International Conference on World Wide Web Companion}, pages 1003--1012.

\bibitem[{Popat et~al.(2018)Popat, Mukherjee, Yates, and Weikum}]{popat2018declare}
Kashyap Popat, Subhabrata Mukherjee, Andrew Yates, and Gerhard Weikum. 2018.
\newblock Declare: Debunking fake news and false claims using evidence-aware deep learning.
\newblock In \emph{Proceedings of the 2018 Conference on Empirical Methods in Natural Language Processing}, pages 22--32.

\bibitem[{Raffel et~al.(2020)Raffel, Shazeer, Roberts, Lee, Narang, Matena, Zhou, Li, Liu et~al.}]{raffel2020exploring}
Colin Raffel, Noam Shazeer, Adam Roberts, Katherine Lee, Sharan Narang, Michael Matena, Yanqi Zhou, Wei Li, Peter~J Liu, et~al. 2020.
\newblock Exploring the limits of transfer learning with a unified text-to-text transformer.
\newblock \emph{J. Mach. Learn. Res.}, 21(140):1--67.

\bibitem[{Reimers and Gurevych(2019)}]{reimers2019sentence}
Nils Reimers and Iryna Gurevych. 2019.
\newblock Sentence-bert: Sentence embeddings using siamese bert-networks.
\newblock In \emph{Proceedings of the 2019 Conference on Empirical Methods in Natural Language Processing and the 9th International Joint Conference on Natural Language Processing (EMNLP-IJCNLP)}, pages 3982--3992.

\bibitem[{Shu et~al.(2019)Shu, Cui, Wang, Lee, and Liu}]{shu2019defend}
Kai Shu, Limeng Cui, Suhang Wang, Dongwon Lee, and Huan Liu. 2019.
\newblock defend: Explainable fake news detection.
\newblock In \emph{Proceedings of the 25th ACM SIGKDD international conference on knowledge discovery \& data mining}, pages 395--405.

\bibitem[{Shu et~al.(2017)Shu, Sliva, Wang, Tang, and Liu}]{shu2017fake}
Kai Shu, Amy Sliva, Suhang Wang, Jiliang Tang, and Huan Liu. 2017.
\newblock Fake news detection on social media: A data mining perspective.
\newblock \emph{ACM SIGKDD explorations newsletter}, 19(1):22--36.

\bibitem[{Silva et~al.(2021)Silva, Han, Luo, Karunasekera, and Leckie}]{silva2021propagation2vec}
Amila Silva, Yi~Han, Ling Luo, Shanika Karunasekera, and Christopher Leckie. 2021.
\newblock Propagation2vec: Embedding partial propagation networks for explainable fake news early detection.
\newblock \emph{Information Processing \& Management}, 58(5):102618.

\bibitem[{Sitaula et~al.(2020)Sitaula, Mohan, Grygiel, Zhou, and Zafarani}]{sitaula2020credibility}
Niraj Sitaula, Chilukuri~K Mohan, Jennifer Grygiel, Xinyi Zhou, and Reza Zafarani. 2020.
\newblock Credibility-based fake news detection.
\newblock In \emph{Disinformation, Misinformation, and Fake News in Social Media}, pages 163--182. Springer.

\bibitem[{Stammbach and Ash(2020)}]{stammbach2020fever}
Dominik Stammbach and Elliott Ash. 2020.
\newblock e-fever: Explanations and summaries for automated fact checking.
\newblock \emph{Proceedings of the 2020 Truth and Trust Online (TTO 2020)}, pages 32--43.

\bibitem[{Stammbach and Neumann(2019)}]{stammbach2019team}
Dominik Stammbach and Guenter Neumann. 2019.
\newblock Team domlin: Exploiting evidence enhancement for the fever shared task.
\newblock In \emph{Proceedings of the Second Workshop on Fact Extraction and VERification (FEVER)}, pages 105--109.

\bibitem[{Szczepa{\'n}ski et~al.(2021)Szczepa{\'n}ski, Pawlicki, Kozik, and Chora{\'s}}]{szczepanski2021new}
Mateusz Szczepa{\'n}ski, Marek Pawlicki, Rafa{\l} Kozik, and Micha{\l} Chora{\'s}. 2021.
\newblock New explainability method for bert-based model in fake news detection.
\newblock \emph{Scientific Reports}, 11(1):1--13.

\bibitem[{Thorne et~al.(2018)Thorne, Vlachos, Christodoulopoulos, and Mittal}]{thorne2018fever}
James Thorne, Andreas Vlachos, Christos Christodoulopoulos, and Arpit Mittal. 2018.
\newblock Fever: a large-scale dataset for fact extraction and verification.
\newblock In \emph{Proceedings of the 2018 Conference of the North American Chapter of the Association for Computational Linguistics: Human Language Technologies, Volume 1 (Long Papers)}, pages 809--819.

\bibitem[{Thorne et~al.(2019)Thorne, Vlachos, Christodoulopoulos, and Mittal}]{thorne2019evaluating}
James Thorne, Andreas Vlachos, Christos Christodoulopoulos, and Arpit Mittal. 2019.
\newblock Evaluating adversarial attacks against multiple fact verification systems.
\newblock Association for Computational Linguistics.

\bibitem[{Vladika and Matthes(2023)}]{vladika2023scientific}
Juraj Vladika and Florian Matthes. 2023.
\newblock Scientific fact-checking: A survey of resources and approaches.
\newblock \emph{arXiv preprint arXiv:2305.16859}.

\bibitem[{Vosoughi et~al.(2018)Vosoughi, Roy, and Aral}]{vosoughi2018spread}
Soroush Vosoughi, Deb Roy, and Sinan Aral. 2018.
\newblock The spread of true and false news online.
\newblock \emph{science}, 359(6380):1146--1151.

\bibitem[{Xu et~al.(2023)Xu, Liu, Wu, and Wang}]{xu-etal-2023-counterfactual}
Weizhi Xu, Qiang Liu, Shu Wu, and Liang Wang. 2023.
\newblock \href {https://doi.org/10.18653/v1/2023.acl-long.374} {Counterfactual debiasing for fact verification}.
\newblock In \emph{Proceedings of the 61st Annual Meeting of the Association for Computational Linguistics (Volume 1: Long Papers)}, pages 6777--6789, Toronto, Canada. Association for Computational Linguistics.

\bibitem[{Yang et~al.(2022)Yang, Vega-Oliveros, Seibt, and Rocha}]{yang2022explainable}
Jing Yang, Didier Vega-Oliveros, Ta{\'\i}s Seibt, and Anderson Rocha. 2022.
\newblock Explainable fact-checking through question answering.
\newblock In \emph{ICASSP 2022-2022 IEEE International Conference on Acoustics, Speech and Signal Processing (ICASSP)}, pages 8952--8956. IEEE.

\bibitem[{Zhang et~al.(2022)Zhang, Zhong, Li, Zhang, and Zhou}]{zhang2022causalrd}
Weifeng Zhang, Ting Zhong, Ce~Li, Kunpeng Zhang, and Fan Zhou. 2022.
\newblock Causalrd: A causal view of rumor detection via eliminating popularity and conformity biases.
\newblock In \emph{IEEE INFOCOM 2022-IEEE Conference on Computer Communications}, pages 1369--1378. IEEE.

\bibitem[{Zhang and Yang(2021)}]{zhang2021survey}
Yu~Zhang and Qiang Yang. 2021.
\newblock A survey on multi-task learning.
\newblock \emph{IEEE Transactions on Knowledge and Data Engineering}, 34(12):5586--5609.

\bibitem[{Zhou et~al.(2020)Zhou, Jain, Phoha, and Zafarani}]{zhou2020fake}
Xinyi Zhou, Atishay Jain, Vir~V Phoha, and Reza Zafarani. 2020.
\newblock Fake news early detection: A theory-driven model.
\newblock \emph{Digital Threats: Research and Practice}, 1(2):1--25.

\bibitem[{Zhou and Zafarani(2019)}]{zhou2019network}
Xinyi Zhou and Reza Zafarani. 2019.
\newblock Network-based fake news detection: A pattern-driven approach.
\newblock \emph{ACM SIGKDD explorations newsletter}, 21(2):48--60.

\bibitem[{Zhou and Zafarani(2020)}]{zhou2020survey}
Xinyi Zhou and Reza Zafarani. 2020.
\newblock A survey of fake news: Fundamental theories, detection methods, and opportunities.
\newblock \emph{ACM Computing Surveys (CSUR)}, 53(5):1--40.

\end{thebibliography}
